\documentclass[journal]{IEEEtran}
\usepackage[font=small,skip=1pt]{caption}
\IEEEoverridecommandlockouts
\usepackage{textcomp}

\usepackage{amsmath} 
\usepackage{amssymb} 
\usepackage{amsfonts} 
\usepackage{graphicx} 
\usepackage{multirow} 
\usepackage{booktabs} 
\usepackage{algorithm} 
\usepackage{algpseudocode} 
\usepackage{cite} 
\usepackage{xcolor} 
\usepackage{float} 
\usepackage{hyperref} 
\usepackage{amsthm} 
\usepackage{bm} 
\usepackage{enumitem} 
\usepackage{inputenc} 
\usepackage[english]{babel} 
\usepackage{lineno} 
\usepackage{mathdots} 
\usepackage{tabularx} 
\usepackage{threeparttable} 
\usepackage{epstopdf} 
\usepackage{fontenc} 
\usepackage{pifont} 
\usepackage{srcltx} 
\usepackage{subfig} 
\usepackage{makecell}

\begin{document}

\title{Neural Network-based Study for Rice Leaf Disease Recognition and Classification: A Comparative Analysis Between Feature-based Model and Direct Imaging Model}

\author{{Farida Siddiqi Prity*, Mirza Raquib*, Saydul Akbar Murad, Md. Jubayar Alam Rafi, Md. Khairul Bashar Bhuiyan and Anupam Kumar Bairagi}

\thanks{ Farida Siddiqi Prity and Mirza Raquib contribute equally. F. S. Prity is from Department of Computer Science \& Engineering, Netrokona University,
Bangladesh. M. Raquib is from Department of Computer and Communication Engineering,
International Islamic University Chittagong, Bangladesh. S. A. Murad is from the School of Computing Sciences and Computer Engineering, University of Southern Mississippi, Hattiesburg, USA. MD. J. A. Rafi is from Department of CSE, Daffodil International University, Bangladesh. MD. K. B. Bhuiyan is from Department of Electrical \& Electronic Engineering Brac University,
Dhaka 1212, Bangladesh . A. K. Bairagi is from Computer Science and Engineering Discipline Khulna University,
Khulna 9208, Bangladesh. 
}}


\maketitle

\begin{abstract}
Rice leaf diseases significantly reduce productivity and cause economic losses, highlighting the need for early detection to enable effective management and improve yields. This study proposes Artificial Neural Network (ANN)-based image-processing techniques for timely classification and recognition of rice diseases. Despite the prevailing approach of directly inputting images of rice leaves into ANNs, there is a noticeable absence of thorough comparative analysis between the Feature Analysis Detection Model (FADM) and Direct Image-Centric Detection Model (DICDM), specifically when it comes to evaluating the effectiveness of Feature Extraction Algorithms (FEAs). Hence, this research presents initial experiments on the Feature Analysis Detection Model, utilizing various image Feature Extraction Algorithms, Dimensionality Reduction Algorithms (DRAs), Feature Selection Algorithms (FSAs), and Extreme Learning Machine (ELM). The experiments are carried out on datasets encompassing bacterial leaf blight, brown spot, leaf blast, leaf scald, Sheath blight rot, and healthy leaf, utilizing 10-fold Cross-Validation method. A Direct Image-Centric Detection Model is established without the utilization of any FEA, and the evaluation of classification performance relies on different metrics. Ultimately, an exhaustive contrast is performed between the achievements of the Feature Analysis Detection Model and Direct Image-Centric Detection Model in classifying rice leaf diseases. The results reveal that the highest performance is attained using the Feature Analysis Detection Model. The adoption of the proposed Feature Analysis Detection Model for detecting rice leaf diseases holds excellent potential for improving crop health, minimizing yield losses, and enhancing overall productivity and sustainability of rice farming.
\end{abstract}


\section{Introduction}
Rice is one of the most vital staple crops worldwide, providing sustenance for over 3.5 billion people, particularly in Asia, Africa, and Latin America \cite{bari2021real}. In Asia alone, where rice forms the backbone of diets, it accounts for over 90\% of global production and consumption \cite{christou2004potential}. This crop contributes to the primary caloric intake for 60\% of the Asian population, underscoring its irreplaceable role in regional food security \cite{dordas2008role}. Globally, rice provides approximately 20\% of human dietary energy and is an essential source of livelihood for millions of smallholder farmers \cite{daniya2022deep, m1}.

Despite its critical importance, rice production faces significant threats from various leaf diseases, which collectively result in severe crop losses annually \cite{sharma2022big,m2}. These diseases, including bacterial blight, rice blast, brown spot, and tungro, can cause staggering yield losses ~\cite{liu2023analysis,liang2019rice}. For example, rice blast alone is known to destroy between 10-30\% of the global rice crop each year, contributing to annual economic losses of approximately \$5 billion ~\cite{jiang2023rice,m3}. Overall, rice leaf diseases contribute to an estimated 15-30\% reduction in global rice yields, equating to millions of tons of rice lost and billions in economic damage each year ~\cite{feng2020investigation}. Such losses have dire implications for food security, affordability, and accessibility in rice-dependent regions, particularly in Asia, where rice diseases threaten the stability of food systems and the livelihoods of vulnerable farming communities ~\cite{ganesan2022hybridization,verma2021prediction}.

The importance of accurately detecting rice leaf diseases cannot be overstated. Early identification enables timely intervention, reducing the need for extensive pesticide applications and limiting the spread of infections to healthy plants ~\cite{quach2023using,al2023using,goluguri2021rice}. In regions highly dependent on rice, such as Asia and Africa, where small-scale farmers rely on high yields to sustain their livelihoods, rapid and precise disease detection is critical  ~\cite{kukana2020hybrid,leonard2014importance}. Timely detection and intervention safeguard yield volumes and enhance the overall quality of rice, reducing reliance on costly agrochemicals that can disrupt ecosystems ~\cite{singh2019sheath,m4}. Furthermore, early disease management helps conserve resources, supports sustainable farming practices, and ultimately strengthens food security at both national and global levels.

Despite advances in digital and AI-based technologies ~\cite{liu2020first,zarbafi2019overview,han2014quantitative,guo2010identification,feng2020investigation} for detecting crop diseases, rice leaf disease detection faces several unresolved challenges. Traditional methods, including manual inspection by experienced agronomists, are constrained by labor intensity, subjectivity, and time requirements, often leading to delayed responses and inaccurate diagnoses ~\cite{barbedo2016review,goralski2020artificial}. Recent studies have applied machine learning techniques like Convolutional Neural Networks (CNNs) and transfer learning for disease identification and classification in rice leaves ~\cite{wu2018development,kujawa2021artificial,escamilla2020applications}. However, these models frequently suffer from limitations, such as reliance on small or imbalanced datasets, which can bias the training process and restrict the model's performance across varied conditions  ~\cite{samborska2014artificial,m6,zorzetto2000processing,smrekar2009development,verma2017optimizing,azim2021effective,yao2009application,islam2019wavelet,ghyar2017computer,saputra2020rice,matin2020efficient,lu2017identification,rahman2020identification,latif2022deep,simhadri2023automatic}. For instance, studies by Rahman et al.~\cite{rahman2020identification} and Latif et al. ~\cite{latif2022deep} utilized small and skewed datasets, impairing the CNN models' generalizability. Moreover, some studies have employed limited feature extraction methods  ~\cite{verma2017optimizing,azim2021effective,yao2009application,islam2019wavelet,ghyar2017computer,saputra2020rice,m5} or neglected image preprocessing steps  ~\cite{ghyar2017computer,rahman2020identification}, potentially reducing the precision and robustness of disease detection. Additionally, many of these approaches~\cite{verma2017optimizing,azim2021effective,yao2009application,islam2019wavelet,ghyar2017computer,saputra2020rice} suffer from overfitting due to a lack of effective Overfitting Reduction Methods (ORMs) like cross-validation or early stopping, further limiting their adaptability in practical applications.

Our study proposes a comprehensive and refined approach to address these limitations. The main contributions of this study are:
\begin{itemize}
\item Employing diverse feature extraction algorithms, including Texture analysis, Grey Level Co-occurrence Matrix (GLCM), Grey Level Difference Matrix (GLDM), Fast Fourier Transform (FFT), and Discrete Wavelet Transform (DWT), to capture complementary features from rice leaf images for robust disease classification.
\item Using data augmentation techniques to expand and balance the dataset, ensuring a representative sample of disease symptoms.
\item Integrating contrast enhancement techniques such as normalization and Adaptive Histogram Equalization to improve the visual clarity of disease symptoms.
\item Incorporating cross-validation and early stopping techniques to mitigate overfitting and enhance model stability across datasets.
\item Applying Dimensionality Reduction Algorithms such as Principal Component Analysis (PCA), Kernel Principal Component Analysis (KPCA), Sparse Autoencoder (Sparse AE), and Stacked Autoencoder (Stacked AE) to refine the feature set and remove irrelevant data.
\item Utilizing Feature Selection Algorithms, including Anova F-measure, Chi-square Test, and Random Tree (RF), to streamline the feature set and improve efficiency.
\item Developing a robust multi-feature extraction model that integrates feature extraction, preprocessing, and optimization techniques significantly enhances detection accuracy and efficiency for rice leaf disease identification.
\end{itemize}
The structure of this paper is as follows: Section \ref{relatedwork} presents a literature review, summarizing prior research on rice leaf disease detection methods. Section \ref{Methodology} outlines the methodology, covering data collection, preprocessing, and model training techniques. In Section ~\ref{Result analysis and discussion}, we discuss our findings, with an emphasis on model performance metrics. Finally, Section ~\ref{conclusion} concludes the paper with a summary of results and directions for future research.

\section{Related works}\label{relatedwork}
Recently, many authors have incorporated AI techniques to classify rice disease. Rice disease detection using AI is mainly based on two strategies proposed till now: Feature Analysis Detection Model and Direct Image-Centric Detection Model. The FADM in rice disease detection involves extracting relevant features from the leaf image to represent its content, such as specific patterns, textures, shapes, or other visual attributes essential for identifying and distinguishing the diseases. In contrast, the DICDM bypasses the feature extraction step and directly uses the raw pixel data for analysis and interpretation. While the Feature Analysis Detection Model focuses on capturing specific patterns and attributes, the Direct Image-Centric Detection Model relies on the entire image for classification and identification purposes.

\subsection{Overview of Previous Feature Analysis Detection Model of Rice Disease Recognition}
Verma et al. presented a Feature Analysis Detection Model for rice disease detection, where the hybrid features are extracted using Discrete Cosine Transform (DCT) ~\cite{verma2017optimizing}. The extracted features are then classified using inverse multi-quadrics Radial Basis Function (RBF) and Decision Tree, significantly improving the recognition efficiency from 16.67\% to 83.34\%. Azim et al. developed a model using GLCM and Local Binary Pattern (LBP) as textural feature descriptors to detect diseases in rice ~\cite{azim2021effective}. XG Boost and Support Vector Machine (SVM) achieved an accuracy of 86.58\% for disease classification using this approach. Yao et al. utilized GLCM and SVM to detect and classify rice diseases~\cite{yao2009application}. Islam et al. employed DWT for multi-resolution analysis of rice disease images, followed by classification using an ensemble of linear classifiers with the Random Subspace Method (RSM)~\cite{islam2019wavelet}. Ghyar et al. utilized GLCM to classify rice diseases ~\cite{ghyar2017computer}. The classification was performed using SVM and Artificial Neural Networks. Saputra et al.~\cite{saputra2020rice} proposed using GLCM as a feature extraction method for text analysis in classifying rice leaf disease images. The classification was done using the K-Nearest Neighbor (KNN) algorithm, which achieved an accuracy of 65.83\%. Table ~\ref{Tab1} represents the summarized findings of various related works focused on rice disease detection, utilizing Feature Analysis Detection Models to highlight key methods, results, and limitations.

\begin{table*}[!h]
\caption{Summary of Related Works for Rice Disease Detection Using Feature Analysis Detection Model}
\centering
\begin{tabular}{|p{1cm}|p{2.5cm}|p{3cm}|p{2cm}|p{3.5cm}|}
\hline
\textbf{Paper} & \textbf{Dataset} & \textbf{Description of Related Works} & \textbf{Results} & \textbf{Limitation} \\ 
\hline
\cite{verma2017optimizing} & Six diseases: 180 images & DCT,RBF, Decision Tree & Accuracy 83.34\% & Dataset is tiny. Use only one FEA. No DRA. No FSA. No ORM. \\ 
\hline
\cite{azim2021effective} & Three diseases: 120 images & GLCM, LBP, XG Boost, SVM & Accuracy 86.58\% & Dataset is tiny. Use only one FEA. Focus on only four diseases. No DRA. No FSA. No ORM. \\ 
\hline
\cite{yao2009application} & Three diseases: 216 images & GLCM, SVM & Accuracy 97.2\% & Dataset is tiny. Use only one FEA. Focus on only three diseases. No DRA. No FSA. No ORM. \\ 
\hline
\cite{islam2019wavelet} & Five diseases: 135 images & DWT, RSM & Accuracy 95\% & Dataset is tiny. Use only one FEA. Focus on only three diseases. No DRA. No FSA. No ORM. \\ 
\hline
\cite{ghyar2017computer} & Two diseases: 80 images & GLCM, Genetic Algorithm, SVM, ANN & Accuracy 92.50\% & Dataset is tiny. Use only one FEA. No image contrast method. No DRA. No ORM. \\ 
\hline
\cite{saputra2020rice}  & Three diseases: 120 images & GLCM, KNN & Accuracy 65.83\% & Dataset is tiny. Use only one FEA. Focus on only three diseases. No DRA. No ORM. \\ 
\hline
\end{tabular}
\label{Tab1}
\end{table*}

\subsection{Overview of Previous Direct Image-Centric Detection Model of Rice Disease Recognition}
Martin et al. applied the AlexNet technique to identify three prevalent rice diseases, achieving an impressive accuracy of 99\% ~\cite{matin2020efficient}. Lu et al. introduced a novel method for rice disease identification based on deep Convolutional Neural Networks (CNNs) ~\cite{lu2017identification}. Rahman et al. developed a CNN model for the classification of eight categories of rice leaf diseases~\cite{rahman2020identification}. Latif et al. introduced a Deep Convolutional Neural Network (DCNN) transfer learning-based approach to accurately detect and classify rice leaf disease ~\cite{latif2022deep}. Simhadri et al. employed a transfer learning approach utilizing 15 pre-trained CNN models to identify rice leaf diseases automatically ~\cite{simhadri2023automatic}. Table ~\ref{Tab2} represents the summarized findings of various related works focused on rice disease detection, utilizing Direct Image-Centric Detection Models to outline key methods, results, and limitations.

\begin{table*}[!h]
\caption{Summary of Related Works for Rice Disease Detection Using Direct Image-Centric Detection Model}
\centering
\begin{tabular}{|p{1cm}|p{1.5cm}|p{3cm}|p{2.5cm}|p{3cm}|}
\hline
\textbf{Paper} & \textbf{Dataset} & \textbf{Description of Related Works} & \textbf{Results} & \textbf{Limitation} \\ 
\hline
\cite{matin2020efficient} & Three diseases: 120 images & AlexNet & Accuracy 99\% & Dataset is tiny. Focus on only three classes. No DRA. No FSA. \\ 
\hline
\cite{lu2017identification} & Two types of images: 500 images & CNN & Accuracy 95.48\% & Dataset is tiny. Focus on only two classes. No DRA. No FSA. No ORM. \\ 
\hline
\cite{rahman2020identification}  & Nine diseases: 1426 images & CNN & Accuracy 93.3\% & Dataset is not balanced. No image contrast method. No DRA. No FSA. \\ 
\hline
\cite{latif2022deep} & Six diseases: 2167 images & CNN & Accuracy 96.08\%, Precision 0.9620, Recall 0.9617, Specificity 0.9921, F1-score 0.9616 & Dataset is not balanced. No DRA. No FSA. \\ 
\hline
\cite{simhadri2023automatic} & Nine diseases: 9074 images & ResNet50, ResNet101, GoogleNet, Shufflenet, MobileNetV2, Efficientnetb0, DenseNet201, AlexNet, Squeezenet, Darknet53, InceptionV3, InceptionResnetV2, Xception & Accuracy 99.64\%, Precision 98.23\%, Recall 98.21\%, F1-Score 98.20\%, Specificity 99.80\% & Dataset is not balanced. No DRA. No FSA. \\ 
\hline
\end{tabular}
\label{Tab2}
\end{table*}

\section{Methodology}\label{Methodology}
\begin{figure*}[h]
\centerline{\includegraphics[width=0.8\linewidth]{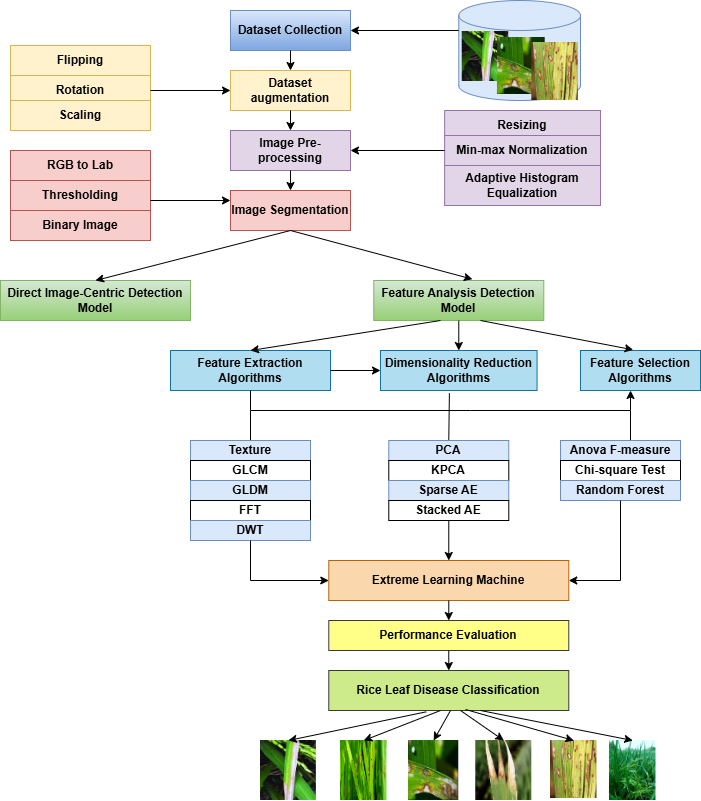}}
\caption{Fundamental process of the proposed study}
\label{fig1}
\end{figure*}

This study aims to develop an ANN model for predicting rice diseases. The proposed system comprised different key stages, as illustrated in Fig.~\ref{fig1} dataset collection, data augmentation, image pre-processing, image segmentation, feature extraction, dimension reduction, feature selection, and classification using ANNs. This section labels the paces and strategies employed for segmenting, extracting features, reducing dimensionality, selecting significant features, and recognizing and classifying the diseases of rice plants used in the proposed system.

\subsection{Dataset collection}
\begin{figure*}[h]
\centerline{\includegraphics[width=0.8\linewidth]{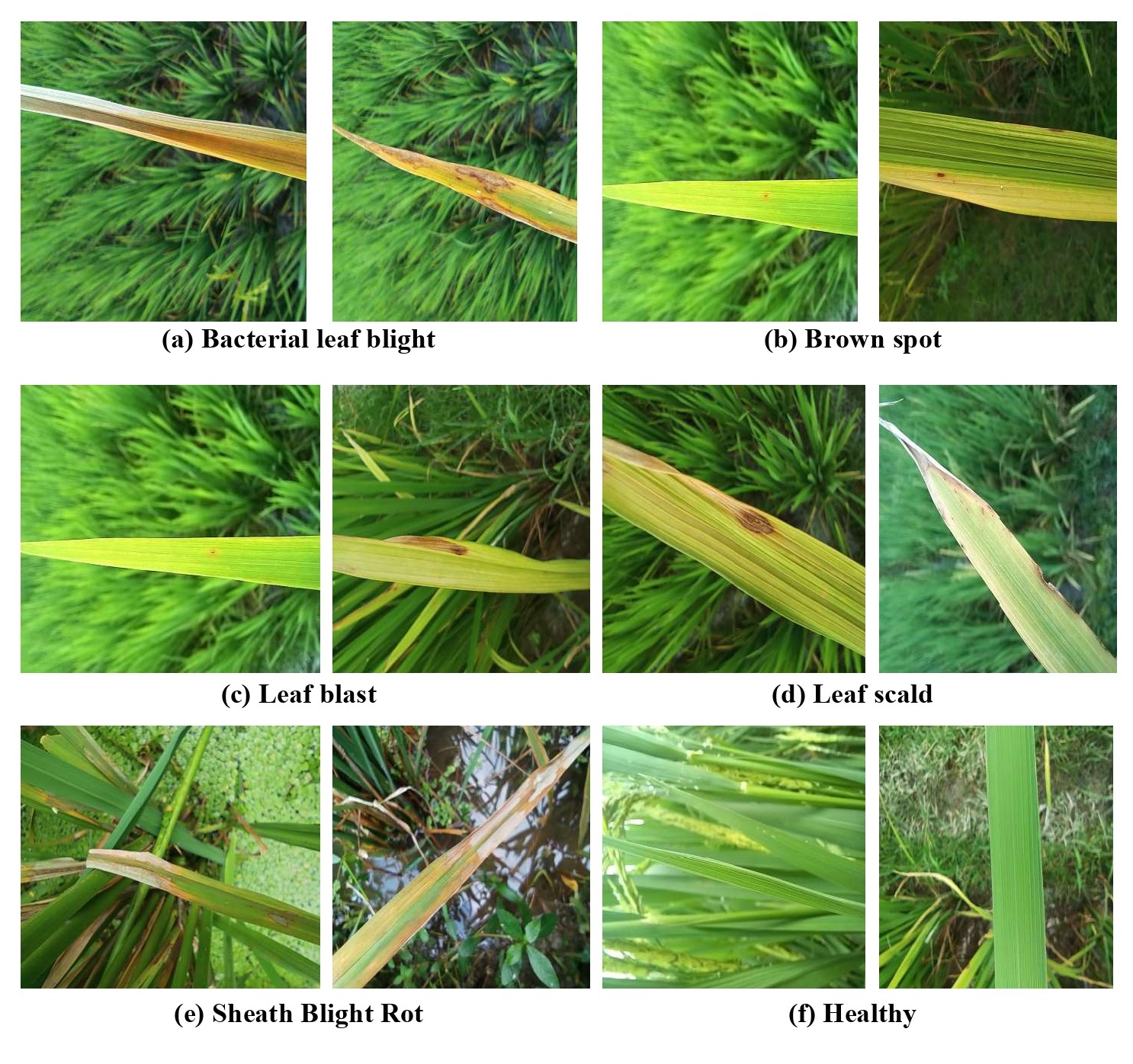}}
\caption{Dataset samples}
\label{fig2}
\end{figure*}

In this research, we exclusively utilized rice leaf disease images sourced from the Kaggle dataset~\cite{rice_leaf_dataset}. The dataset contains images classified into six categories: five representing distinct diseases—bacterial leaf blight (636 images), brown spot (646 images), leaf blast (634 images), leaf scald (628 images), and sheath blight rot (632 images)—and one representing healthy leaves (653 images). Fig. ~\ref{fig2} illustrates representative images from each category, highlighting the characteristic symptoms of the diseases and the normal features of healthy leaves.

\subsection{Dataset augmentation}
To enhance the dataset’s diversity and strengthen model robustness, we applied data augmentation techniques, including flipping, rotation, and scaling, to introduce variations in orientation and size. Horizontal flipping was implemented by using:

\begin{equation}
f(x, y) = f(W - x, y) \tag{1}
\end{equation}

where, \( W \) represents the image width, and each pixel at position \( (x, y) \) is mirrored along the horizontal axis. Rotation was performed by an angle \( \theta \), represented by the transformation in:
\begin{equation}
\begin{pmatrix}
x' \\ 
y'
\end{pmatrix}
=
\begin{pmatrix}
\cos\theta & \sin\theta \\ 
-\sin\theta & \cos\theta
\end{pmatrix}
\begin{pmatrix}
x \\ 
y
\end{pmatrix} \tag{2}
\end{equation}

where, \( (x, y) \) are the original coordinates and \( (x', y') \) are the rotated coordinates. Scaling was applied using the transformation shown as follows:

\begin{equation}
\begin{pmatrix}
x' \\ 
y'
\end{pmatrix}
= s
\begin{pmatrix}
x \\ 
y
\end{pmatrix} \tag{3}
\end{equation}

where, \( s \) denotes the scaling factor applied to each coordinate. Through these augmentation techniques, we increased the dataset to 1000 images per class, yielding a more comprehensive range of samples and enhancing the model’s capacity to generalize effectively across various conditions.

\subsection{Image pre-processing}
The images were transformed to a standardized dimension of \( 256 \times 256 \) pixels to ensure consistency across the dataset. Min-max normalization was applied to scale pixel intensity values, enhancing uniformity and coherence ~\cite{narin2021automatic}. This normalization process is defined by:

\begin{equation}
I_{\text{norm}} = \frac{I - I_{\text{min}}}{I_{\text{max}} - I_{\text{min}}} \tag{4}
\end{equation}

where, \( I \) represents the original pixel intensity, \( I_{\text{min}} \) and \( I_{\text{max}} \) are the minimum and maximum pixel values, respectively, and \( I_{\text{norm}} \) denotes the normalized pixel intensity. Additionally, Adaptive Histogram Equalization (AHE) ~\cite{chen2020epidemiological} was implemented to improve local contrast by adjusting the brightness in each image. AHE enhances the visibility of finer details by locally modifying the histogram for sub-regions within the image, effectively amplifying brightness and improving feature clarity.

\subsection{Image segmentation}
Image segmentation is crucial in accurately identifying a leaf image's diseased portion. The Red Green Blue (RGB) image is initially transformed into the Lab color space, representing color by three distinct values: \( L^* \), \( a^* \), and \( b^* \). This transformation can be mathematically represented using:

\[
\begin{bmatrix}
L^* \\ 
a^* \\ 
b^*
\end{bmatrix}
= f
\begin{bmatrix}
R \\ 
G \\ 
B
\end{bmatrix}
\tag{5}
\]

where, \( f \) denotes the conversion function that translates the RGB color values to the Lab color space. Among these, the \( a^* \) component captures the range from green to red, making it essential for segmenting the affected area. By isolating the \( a^* \) component, the image undergoes a global thresholding process to convert it into a binary image \cite{bijoy2024towards,pizer1987adaptive}. This thresholding can be defined as follows:

\[
I_{\text{binary}}(x,y) =
\begin{cases} 
1 & \text{if } a^*(x,y) > T \\
0 & \text{if } a^*(x,y) \leq T
\end{cases}
\tag{6}
\]

where, \( T \) is the threshold value, and \( I_{\text{binary}}(x,y) \) represents the binary output for each pixel \((x,y)\). The resulting binary image is then overlaid onto the original RGB image, effectively delineating the diseased regions with enhanced clarity.

\subsection{Direct Image-Centric Detection Model}
In the Direct Image-Centric Detection Model, segmented images are fed directly into the Artificial Neural Network (ANN) for classification. Each image, resized to \(256 \times 256\) pixels, serves as the input to the ANN, resulting in 65,536 input nodes per training instance. This direct input approach leverages the pixel intensity values from each segmented image, enabling the ANN to analyze and classify disease patterns directly, bypassing the need for prior feature extraction.

\subsection{Feature Analysis Detection Model}
The Feature Analysis Detection Model utilizes feature extraction to identify and classify disease patterns in images through a structured, multi-stage process. This model is organized into three key stages:
\begin{itemize}

    \item Feature Extraction Algorithms stage
    \item Dimensionality Reduction Algorithms stage
    \item Feature Selection Algorithms stage
\end{itemize}

\subsubsection{Feature Extraction Algorithms Stage}

Feature Extraction Algorithms (FEAs) are computational techniques in machine learning that convert raw data into meaningful features. These features capture essential data aspects and serve as inputs for models, enhancing pattern learning, prediction, and task performance. The proposed study calculates spatial and frequency domain features for rice disease recognition. This study employs five FEAs to accomplish this task: Texture \cite{he1987texture}, GLCM \cite{de2013multi}, GLDM \cite{sen2009counting}, FFT \cite{duhamel1990fast}, and DWT \cite{nason1994discrete}.

The FEAs utilized in this study included calculating 14 statistical features for each image, capturing essential characteristics. These features include area, mean, standard deviation, energy, median, skewness, entropy, maximum value, minimum value, mean absolute deviation, kurtosis, range, root mean square, and uniformity. These metrics were computed directly from the segmented 256×256 images, providing a foundational dataset to describe variations within the images.

Grey Level Co-occurrence Matrix (GLCM) and Grey Level Difference Matrix (GLDM) features were computed across four orientations: \(0^\circ\), \(45^\circ\), \(90^\circ\), and \(135^\circ\). For each orientation, 14 features were calculated, yielding a total of 56 features per method (\(4 \text{ orientations} \times 14 \text{ features}\)). These features capture textural properties, contributing to the model’s ability to distinguish diseased regions based on directional texture patterns. For instance, the GLCM for an image \(f(x, y)\) is computed as follows:

\[
p(i, j) = \frac{1}{N} \sum_{x=1}^{M} \sum_{y=1}^{M} \delta(f(x, y) = i, f(x+dx, y+dy) = j)
\tag{7}
\]

where \(p(i, j)\) represents the probability of pixel intensity \(i\) co-occurring with intensity \(j\) at a specific displacement vector \((dx, dy)\), and \(M\) is the total number of gray levels in the image.

The Discrete Wavelet Transform (DWT) decomposes each image into eight distinct sub-bands, representing various frequency components within the image. For each sub-band, 14 features were calculated, resulting in a total of 112 features from DWT (\(8 \text{ bands} \times 14 \text{ features}\)). DWT is expressed mathematically as follows:

\[
W_\phi(a, b) = \frac{1}{\sqrt{a}} \int_{-\infty}^{\infty} f(t) \phi\left(\frac{t-b}{a}\right) dt
\tag{8}
\]

where \(W_\phi(a, b)\) represents the wavelet coefficient, \(a\) is the scaling factor, \(b\) is the translation factor, and \(\phi\) is the wavelet function.

Fast Fourier Transform (FFT) was applied to analyze the frequency domain by converting spatial data into frequency components. The FFT transformation is represented as:

\[
F(u, v) = \sum_{x=0}^{M-1} \sum_{y=0}^{N-1} f(x, y) e^{-2\pi i \left(\frac{ux}{M} + \frac{vy}{N}\right)}
\tag{9}
\]

where \((x, y)\) is the pixel intensity at spatial coordinates, and \((u, v)\) is the frequency component at frequency coordinates \((u, v)\).

Texture features were also directly calculated from the segmented images. In total, each image in the dataset contributed a comprehensive set of 252 features: 14 texture features, 56 GLCM features, 56 GLDM features, and 112 DWT features. These features collectively form a robust dataset, encapsulating both spatial and frequency domain information necessary for effective disease recognition and classification.

In the Feature Extraction Algorithms stage, eight distinct phases assess the classification performance of various feature sets, each using an Artificial Neural Network (ANN) for classification. First, 14 texture features are extracted and classified by the ANN, followed by 56 GLCM features and 56 GLDM features, each classified using the ANN. Additionally, 14 FFT features and 14 DWT features are separately classified. A combined feature vector, termed “All,” integrates all 252 features from Texture, FFT, GLCM, DWT, and GLDM and is subsequently classified by the ANN. Moreover, a subset of 126 frequency domain features (comprising FFT and DWT features) and another subset of 126 spatial domain features (combining Texture, GLCM, and GLDM features) are classified individually using the ANN. This approach allows the model to evaluate both isolated and combined feature sets, facilitating a robust assessment of each feature’s effectiveness in accurate disease classification.

\subsubsection{Dimensionality Reduction Algorithms Stage}

Dimensionality Reduction Algorithms simplify datasets by reducing the number of features while retaining relevant information. These algorithms transform data into a lower-dimensional representation, enhancing efficiency and minimizing overfitting risks. This study employs four Dimensionality Reduction Algorithms: PCA \cite{abdi2010principal}, KPCA \cite{scholkopf1997kernel}, Sparse AE \cite{makhzani2013k}, and Stacked AE \cite{zabalza2016novel}.

PCA reduces the dimensionality of the feature vector by identifying directions (principal components) that maximize variance in the data. Given a feature vector \(\mathbf{X} = [x_1, x_2, x_3, \dots, x_n]\), the PCA transformation is represented by:
\[
\mathbf{X}' = \mathbf{X}\mathbf{W} \tag{10}
\]

where \(\mathbf{W}\) is the matrix of eigenvectors of the covariance matrix of \(\mathbf{X}\). This study applies PCA to reduce the feature vector from 252 to 70 features.

KPCA extends PCA by applying a non-linear kernel function to map data into a higher-dimensional space, then performing linear PCA in that space. The kernel transformation is represented as follows:

\[
K_{ij} = \phi(x_i) \cdot \phi(x_j) \tag{11}
\]

where \(\mathbf{K}\) is the kernel matrix, capturing inner products in the feature space using a non-linear mapping \(\phi\). KPCA reduces the feature vector from 252 to 65 features.

\begin{figure*}[h]
\centerline{\includegraphics[width=0.8\linewidth]{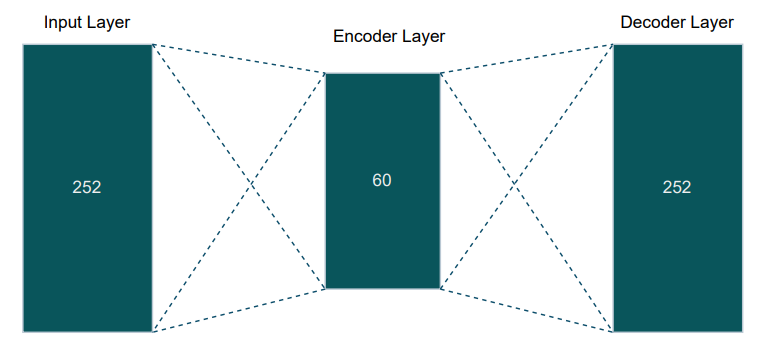}}
\caption{Design of proposed Sparse AE}
\label{fig3}
\end{figure*}

The Sparse Autoencoder algorithm compresses high-dimensional feature vectors into a lower-dimensional space of 60 features. This architecture includes an input layer for the 252 features, an encoder layer that reduces dimensionality to 60 features, and subsequent decoder layers that reconstruct the input from the encoded representation. The loss function for Sparse AE is given by:

\[
L = \| \mathbf{X} - \mathbf{X}' \|^2 + \gamma \sum_{j=1}^m \text{KL}(\rho \| p_j) \tag{12}
\]

where \(\| \mathbf{X} - \mathbf{X}' \|^2\) is the reconstruction error, \(\gamma\) is the sparsity penalty, \(\text{KL}\) denotes the Kullback-Leibler divergence, \(\rho\) is the desired sparsity, and \(p_j\) is the average activation of hidden unit \(j\). The specific design of the Sparse AE employed is detailed in Fig.~\ref{fig3}.

\begin{figure*}[h]
\centerline{\includegraphics[width=0.8\linewidth]{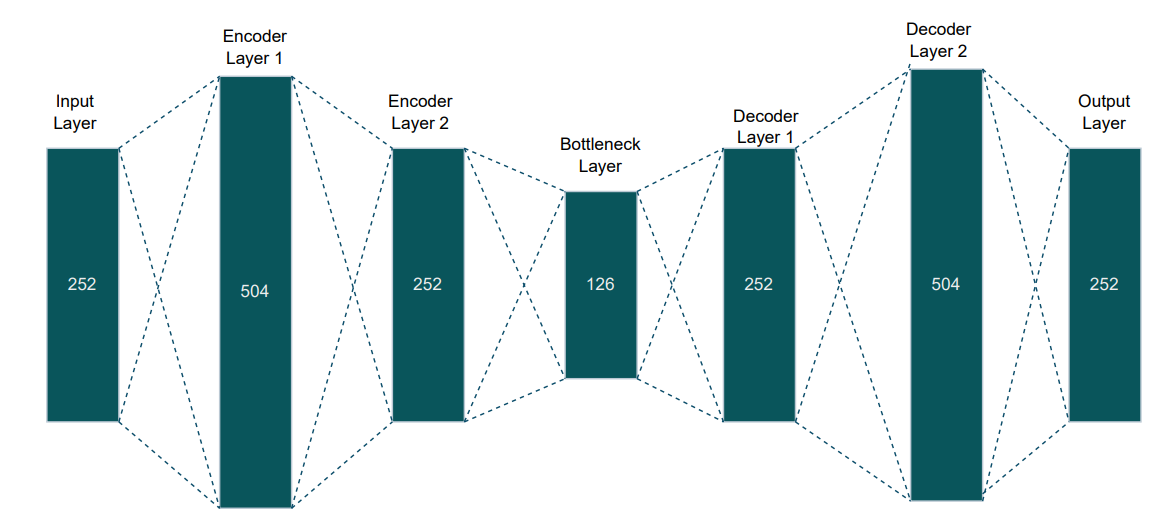}}
\caption{Design of proposed Stacked AE}
\label{fig4}
\end{figure*}

The Stacked Autoencoder structure used in this study reduces the dimensionality of the vector from 252 features to 126 features. This architecture includes an input layer with 252 features, followed by two encoder layers that successively compress the data. The output of the second encoder layer serves as the bottleneck, yielding the final feature vector of 126 features. Decoder layers then reconstruct the input data from this bottleneck representation. A detailed depiction of the Stacked Autoencoder’s architecture, illustrating the progression of information across layers, is presented in Fig. ~\ref{fig4}. Each encoder layer performs the transformation as follows:

\[
\mathbf{H} = f(\mathbf{W}\mathbf{X} + \mathbf{b}) \tag{13}
\]

where \(\mathbf{W}\) is the weight matrix, \(\mathbf{b}\) is the bias, and \(f\) is the activation function. These reduced feature sets from each DRA are subsequently fed into the Artificial Neural Network for classification.

\subsubsection{Feature Selection Algorithms Stage}

Feature Selection Algorithms (FSAs) are computational techniques used in AI and data analysis to identify and select the most relevant features from a larger set of inputs. Feature selection aims to improve model performance by reducing noise, overfitting, and computational complexity. This study employs three Feature Selection Algorithms: Anova F-measure \cite{archer1997sensitivity}, Chi-square Test \cite{tallarida1987chi}, and Random Forest (RF) \cite{belgiu2016random}.
  
The Anova F-measure assesses the variance between groups to identify features with strong discriminative power. The F-score for each feature is calculated as follows:

\[
F = \frac{\text{Between-group variance}}{\text{Within-group variance}} \tag{14}
\]

In this phase, the Anova F-measure identifies 50 out of the original 252 features as significantly discriminative. The Chi-square test evaluates the independence between categorical variables and each feature, calculating how expected frequencies deviate from observed values. The Chi-square score for each feature is computed as follows:

\[
\chi^2 = \sum_{i=1}^k \frac{(O_i - E_i)^2}{E_i} \tag{15}
\]

where \(O_i\) and \(E_i\) are the observed and expected frequencies, respectively, for each category \(i\). This test identifies 40 significant features from the initial set of 252. 

Random Forest selects features based on their importance in the context of tree-based modeling. The importance score of a feature \(x_i\) is calculated by observing the mean decrease in impurity (MDI) across all trees in the forest:

\[
\text{Importance}(x_i) = \frac{\sum_{t=1}^T \Delta I_t(x_i)}{T} \tag{16}
\]

where \(T\) is the total number of trees, and \(\Delta I_t(x_i)\) is the decrease in impurity in tree \(t\) due to feature \(x_i\). Using RF, 35 essential features are selected from the original 252. These selected features from each FSA are subsequently fed into the Artificial Neural Network for classification.

\subsection{Artificial Neural Network}

\begin{figure*}[h]
\centerline{\includegraphics[width=0.8\linewidth]{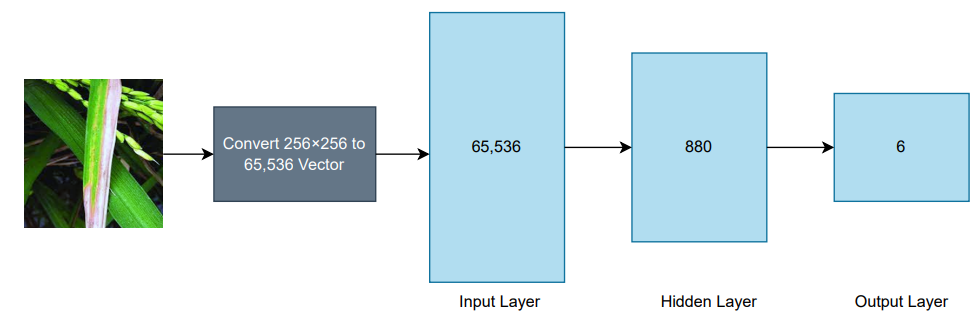}}
\caption{Architecture of ELM for Direct Image-Centric Detection Model}
\label{fig5}
\end{figure*}

Artificial Neural Networks (ANNs) are a powerful tool in modern artificial intelligence, enabling computers to learn and make predictions based on data by emulating, to a certain extent, the information processing mechanisms of the human brain. Their capacity to identify complex patterns and relationships has contributed significantly to advancements across numerous applications, establishing ANNs as a foundational technology in today’s technological landscape \cite{gupta1997artificial}. This research specifically employs the Extreme Learning Machine (ELM) neural network to classify rice leaf diseases. In this model, segmented images are directly inputted into the ELM, with each \(256 \times 256\) pixel image providing 65,536 input features per training instance. The hidden layer of the ELM is configured with 880 neurons, while a single output classifier categorizes rice disease images into six classes. The architecture of the ELM for this model is presented in Fig.~\ref{fig5}.

For the Feature Analysis Detection Model, various configurations of hidden layer neurons in the Extreme Learning Machine (ELM) have been tested. Optimal classification accuracy is achieved when the number of hidden neurons \( L \) is set to twice the number of input features, in accordance with the Feature Extraction Algorithms (FEAs), Dimensionality Reduction Algorithms (DRAs), and Feature Selection Algorithms (FSAs) stages. This configuration uses a single output classifier to categorize rice disease images into six classes. The ELM’s output \( O \) is computed as follows:

\[
O = (H \times \beta)\tag{17} 
\]

where:
\begin{itemize}
    \item \( H \) represents the hidden layer output matrix.
    \item \( \beta \) denotes the output weight matrix.
    \item \( g \) is the activation function applied to the output layer.
\end{itemize}

\begin{figure*}[h]
\centerline{\includegraphics[width=0.8\linewidth]{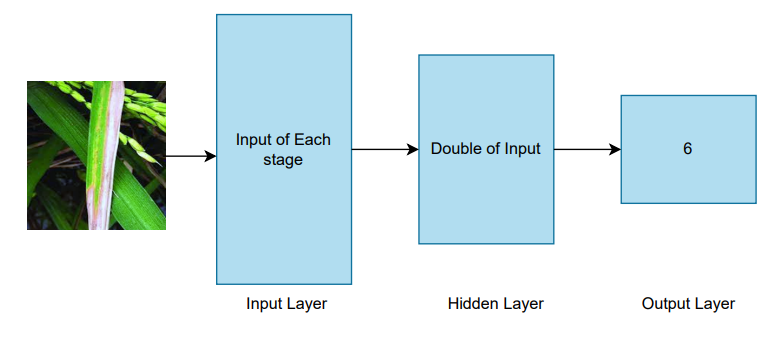}}
\caption{Architecture of ELM for Feature Analysis Detection Model}
\label{fig6}
\end{figure*}

The architecture of the ELM for the Feature Analysis Detection Model is illustrated in Fig.~\ref{fig6}. Early stopping and 10-fold cross-validation have been implemented to mitigate overfitting. The hyperparameters for the learning algorithm are specified in Table~\ref{Tab3}.

\begin{table}[!h]
\centering
\caption{Hyper-parameters of the ELM}
\begin{tabular}{|l|l|}
\hline
\textbf{Hyper-parameters} & \textbf{Values} \\ \hline
Loss function            & Categorical cross-entropy \\ \hline
Optimizer                & Adam \\ \hline
Activation Function      & Softmax \\ \hline
Maximum epochs          & 100 \\ \hline
Learning rate           & 0.001 \\ \hline
Train ratio             & 0.8 \\ \hline
Test ratio              & 0.2 \\ \hline
Batch size              & 16 \\ \hline
\end{tabular}
\label{Tab3}
\end{table}

\subsection{Performance Evaluation}

This research utilizes model creation to assess the effectiveness and utility of Artificial Neural Networks (ANNs) in predicting rice diseases. The performance of the models is evaluated using the following metrics: sensitivity, specificity, precision, F-measure, and accuracy.

Sensitivity is the ability of the model to correctly identify actual positives, measuring the true positive rate. It is calculated using the following equation:
\[
\text{Sensitivity} = \frac{TP}{TP + FN} \tag{18} 
\]
where \(TP\) represents the number of true positives and \(FN\) represents the number of false negatives.

Specificity is the ability of the model to correctly identify actual negatives, indicating the true negative rate. It can be calculated using the following equation:
\[
\text{Specificity} = \frac{TN}{TN + FP} \tag{19} 
\]
where \(TN\) represents the number of true negatives and \(FP\) represents the number of false positives.

Precision calculates the proportion of correctly identified positives among all predicted positives, showing the prediction accuracy for positives. The formula for precision is as follows:
\[
\text{Precision} = \frac{TP}{TP + FP} \tag{20} 
\]
where \(TP\) and \(FP\) are the counts of true positives and false positives, respectively.

F-measure is the harmonic mean of precision and sensitivity, balancing both for a combined performance score. It can be calculated using the following equation:
\[
\text{F-measure} = \frac{2 \times \text{Precision} \times \text{Sensitivity}}{\text{Precision} + \text{Sensitivity}} \tag{21} 
\]

Accuracy is the overall correctness of the model, representing the proportion of true results (both positives and negatives) among total cases. It is calculated using the following equation:
\[
\text{Accuracy} = \frac{TP + TN}{TP + TN + FP + FN} \tag{22} 
\]
where \(TP\), \(TN\), \(FP\), and \(FN\) represent the counts of true positives, true negatives, false positives, and false negatives, respectively.

\section{Result analysis and discussion}\label{Result analysis and discussion}
Rice is a vital staple crop, and diseases affecting rice can origin substantial damage in harvest, posing threats to food security. Timely disease detection allows for swift intervention, minimizing the spread and severity of diseases. Accurate identification enables the implementation of site-specific treatments and interventions tailored to the specific disease, reducing the need for excessive chemical usage and promoting sustainable farming practices. Rice disease detection using ANNs has gained extensive consideration because of its potential to revolutionize disease management in rice crops. So this study has projected an automated ANN-based system to recognize rice leaf disease and diminish the food security threat.
\subsection{Performance Evaluation }
In this sub-segment, the classification outcomes of both the Feature Analysis Detection Model and Direct Image-Centric Detection Model (comprising FEAs stage, DRAs stage, and FSAs stage) in recognizing and categorizing rice leaf diseases are presented. A variety of performance metrics, including accuracy, sensitivity, specificity, precision, and F-measure, are employed to gauge the efficacy of these models. In the DICDM, the segmented images are inputted straight into the ELM for classification. The FADM is structured around three primary stages: FEAs, DRAs, and FSAs. During the FEAs stage, diverse FEAs are applied to the segmented images to extract pertinent features, which are subsequently utilized for ELM-based classification. In the DRAs stage, a range of techniques—such as PCA, KPCA, Sparse AE, and Stacked AE—are employed to streamline the feature. The objective is to condense the feature vector's dimensionality, leading to a refined feature vector that is then employed for ELM classification. In the FSAs stage, the focus shifts to algorithms like Anova F-measure, Chi-Square Test, and RF. These algorithms sift through the 252-feature feature sets, identifying the most crucial features for ELM-based classification. The classification results from the DIM, FEAs stage, DRAs stage, and FSAs stage are comprehensively summarized in Table ~\ref{Tab4}.

The DCIM utilizing the ELM demonstrates the following classification metrics for model construction: a sensitivity of 73\%, specificity of 77\%, precision of 65\%, F-measure of 64\%, and accuracy of (74.97 ± 0.8) \%. The FADM is structured around three pivotal stages: FEAs, DRAs, and FSAs. In the FEAs stage, the amalgamation of all FEAs (referred to as "All") yields the highest accuracy. The "All" features model achieves notable metrics, including a sensitivity of 95\%, specificity of 96\%, precision of 93\%, F-measure of 94\%, and an accuracy of (94.87 ± 0.6)\%. The DRAs stage encompasses PCA, KPCA, Sparse AE, and Stacked AE. Among these, Kernel Principal Component Analysis attains the highest accuracy of (98.99 ± 0.2) \%. In the FSAs stage, Anova F-measure garners the highest accuracy of (94.31 ± 0.6)\%.

Among the three stages of the FBM, the DRAs stage emerges as the most proficient. Specifically, Kernel Principal Component Analysis demonstrates unparalleled classification prowess. In this phase, KPCA is applied to a feature vector initially comprising 252 features derived from five FEAs. The application of KPCA effectively reduces the feature vector's dimensionality from 252 features to 65 features. By leveraging KPCA, the proposed system achieves remarkable classification performance metrics: a staggering accuracy of (98.99 ± 0.2)\%, along with an impressive 99\% sensitivity, 98\% specificity, 99\% precision, and a commendable F-measure of 97\% across all six classes. Feature Analysis Detection Model and Direct Image-Centric Detection Model.

\begin{table*}[ht]
\centering
\caption{Performance Evaluation}
\resizebox{\textwidth}{!}{
\begin{tabular}{|c|c|c|c|c|c|c|c|}
\hline
\textbf{Model/Stages} & \textbf{Classifier/Algorithms} & \textbf{Pixel Size/Feature Number} & \textbf{Sensitivity (\%)} & \textbf{Specificity (\%)} & \textbf{Precision (\%)} & \textbf{F-measure (\%)} & \textbf{Accuracy (\%)} \\
\hline
\multirow{1}{*}{DICDM} & ELM & 256×256 & 73 & 77 & 65 & 64 & 74.97 ± 0.8 \\
\hline
\multirow{7}{*}{FEAs Stag} & \multicolumn{7}{c|}{FADM} \\
\hline
& Texture & 14 & 55 & 58 & 54 & 57 & 52.21 ± 2.6 \\
& GLCM & 56 & 91 & 94 & 91 & 90 & 90.23 ± 0.9 \\
& GLDM & 56 & 77 & 79 & 80 & 77 & 77.11 ± 0.23 \\
& FFT & 14 & 66 & 68 & 63 & 61 & 66.46 ± 1.5 \\
& DWT & 112 & 88 & 90 & 87 & 90 & 89.41 ± 1.1 \\
& Texture + GLCM + GLDM + FFT + DWT (All) & 252 & 95 & 96 & 93 & 94 & 94.87 ± 0.6 \\
& Frequency Domain (FFT + DWT) & 126 & 89 & 91 & 88 & 87 & 87.98 ± 1.3 \\
& Spatial Domain (Texture + GLCM + GLDM) & 126 & 93 & 95 & 91 & 92 & 92.38 ± 0.8 \\
\hline
\multirow{4}{*}{DRAs Stage} & PCA & 70 & 98 & 99 & 96 & 97 & 97.61 ± 0.3 \\
& KPCA & 65 & 99 & 98 & 99 & 97 & 98.99 ± 0.2 \\
& Sparse Autoencoder & 60 & 42 & 45 & 22 & 26 & 41.88 ± 5.2 \\
& Stacked Autoencoder & 126 & 37 & 35 & 20 & 23 & 35.37 ± 0.2 \\
\hline
\multirow{3}{*}{FSAs Stage} & Anova F-measure & 50 & 95 & 94 & 94 & 93 & 94.31 ± 0.6 \\
& Chi-square Test & 40 & 94 & 93 & 94 & 93 & 93.29 ± 0.7 \\
& Random Forest & 35 & 94 & 95 & 94 & 94 & 93.87 ± 0.7 \\
\hline
\end{tabular}
}
\label{Tab4}
\end{table*}

\subsection{Comparisons of Feature Analysis Detection Model with the Direct Image-Centric Detection Model}
This research primarily emphasizes the importance of feature extraction using the Feature Analysis Detection Model. It investigates three significant stages of the FADM: FEAs stage, DRAs stage, and FSAs stage. A comparison is made with the Direct Image-Centric Detection Model, which excludes FEAs. In the DICDM, images are directly classified with ELM, achieving an accuracy of (74.97 ± 0.8)\%. Among the FADM, the Kernel Principal Component Analysis Dimensionality Reduction Algorithm achieves the highest accuracy of (98.99 ± 0.2)\%.

\begin{figure*}[h]
\centerline{\includegraphics[width=0.8\linewidth]{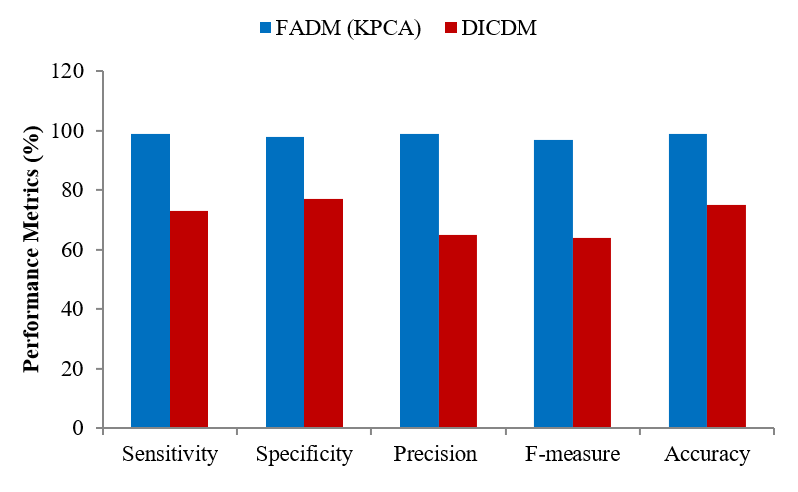}}
\caption{Comparative analysis between FADM and DICDM}
\label{fig7}
\end{figure*}

When comparing the performance of the FADM (KPCA) with the DICDM, it becomes clear that KPCA offers better efficiency. The FADM (KPCA) surpasses the DICDM in terms of performance metrics. By extracting appropriate features, the Feature Analysis Detection Model streamlines machine training, simplifies model development, and accelerates learning and generalization. In contrast, directly inputting images to the classifier presents challenges in learning and training. Thus, the Feature Analysis Detection Model consistently outperforms the Direct Image-Centric Detection Model. The superiority of the applied FADM (KPCA) over the Direct Image-Centric Detection Model is evident from Fig.~\ref{fig7}.

\subsection{Comparisons with the prior research}
This study presents a series of insightful comparative analyses with existing research:	First and foremost, prior investigations ~\cite{sharma2022big,wang2022review} have probed the reliability of classifiers in detecting positive instances of rice diseases. Yet, specificity is a pivotal factor in curbing disease transmission, which precisely identifies healthy plants. This capability facilitates early detection and isolation of diseased plants and enables targeted management strategies, containment of asymptomatic carriers, and disease progression monitoring. In stark contrast to prior works ~\cite{kathiresan2021disease}, our implemented Feature Analysis Detection Model (KPCA) boasts enhanced specificity (98\%), exemplifying a robust barrier against rice disease dissemination.

Secondly, diverging from many AI techniques ~\cite{azim2021effective,yao2009application,ghyar2017computer,saputra2020rice} that fixate solely on the spatial domain, this study underscores features derived from both the spatial and frequency domains through the FEAs.

Thirdly, several studies~\cite{verma2017optimizing,azim2021effective,yao2009application,islam2019wavelet,matin2020efficient,rahman2020identification,latif2022deep} have attained high classification accuracy for rice diseases. However, their models are vulnerable to overfitting due to the absence of overfitting mitigation mechanisms. In contrast, our model employs Early Stopping and Cross-Validation with precision to effectively counter overfitting challenges.

Lastly, to affirm the efficacy of the proposed system, we rigorously compare the performance of our FADM with existing works. The meticulous comparison delineated in Table ~\ref{Tab5}. underscores the superior performance of our implemented Feature Analysis Detection Model when juxtaposed with prior research. 


\begin{table*}[]
\caption{Comparisons between our FADM and existing works}
\centering
\label{Tab5}
\resizebox{\textwidth}{!}{
\begin{tabular}{|c|c|c|c|c|c|}
\hline
\textbf{Model}                                & \textbf{Sensitivity (\%)} & \textbf{Specificity (\%)} & \textbf{Precision (\%)} & \textbf{F-measure (\%)} & \textbf{Accuracy (\%)} \\ \hline
Proposed FADM (KPCA)                          & 99                        & 98                        & 99                      & 97                      & 98.99 ± 0.2            \\ \hline
\cite{latif2022deep}         & 96.17                     & 99.21                     & 96.2                    & 96.16                   & 96.08                  \\ \hline
\cite{daniya2022exponential} & 92.3                      & 91.9                      & -                       & -                       & 91.60                  \\ \hline
\cite{sharma2022big}         & 98.75                     & -                         & 98.81                   & 98.70                   & 98.70                  \\ \hline
\cite{verma2017optimizing}   & -                         & -                         & -                       & -                       & 83.34                  \\ \hline
\cite{yao2009application}    & -                         & -                         & -                       & -                       & 97.2                   \\ \hline
\cite{azim2021effective}     & -                         & -                         & -                       & -                       & 86.58                  \\ \hline
\cite{saputra2020rice}       & -                         & -                         & -                       & -                       & 65.83                  \\ \hline
\end{tabular}
}
\end{table*}

\subsection{Impact of Proposed Feature Analysis Detection Model on Economy}
Rice disease detection using Artificial Neural Networks significantly impacts the economy by increasing crop yield, reducing costs, improving market competitiveness, enhancing food security, and fostering technological advancements. Timely detection and effective management of diseases through ANNs lead to higher rice yields, ensuring a stable supply of rice and contributing to agricultural productivity and economic growth. By optimizing resource allocation and reducing unnecessary inputs, farmers can save costs while maintaining effective disease control. Disease-free rice fetches better prices in the market, improving market competitiveness and profitability. Additionally, adopting ANN-based detection systems drives technological advancements, innovation, and research in agriculture, promoting sustainable practices and improved agricultural outcomes. Overall, rice disease detection using our proposed Feature Analysis Detection Model has wide-ranging economic benefits, supporting food security, enhancing farmer livelihoods, and driving economic development in rice-dependent regions.

\section{Conclusion}\label{conclusion}
Rice diseases significantly impact crop yield and global food security, necessitating advanced, accurate disease management strategies. This study proposes a neural network-based framework for automated recognition and classification of rice diseases using leaf imagery. Feature extraction techniques such as Texture analysis, GLCM, GLDM, FFT, and DWT extract critical image characteristics, while dimensionality reduction (PCA, KPCA, Sparse AE, Stacked AE) and feature selection (Anova F-measure, Chi-square Test, RF) optimize classification. The framework uses Extreme Learning Machine (ELM) to categorize six classes: bacterial leaf blight, brown spot, leaf blast, leaf scald, sheath blight rot, and healthy leaves, achieving early and precise disease identification. The FADM model demonstrates exceptional performance, achieving 98.99 ± 0.2\% accuracy, 99\% sensitivity, and 98\% specificity, largely attributed to KPCA. Despite its success, the study's modest sample size highlights the need for future research with larger datasets.

Future directions include extending the methodology to classify diseases in other crops like potatoes and maize. Integrating this approach into agricultural practices can optimize resources, reduce yield losses, and promote sustainable cultivation, ensuring global food security while benefiting farmers and consumers alike.

\bibliographystyle{IEEEtran}

\begin{thebibliography}{100}
\providecommand{\url}[1]{#1}
\csname url@samestyle\endcsname
\providecommand{\newblock}{\relax}
\providecommand{\bibinfo}[2]{#2}
\providecommand{\BIBentrySTDinterwordspacing}{\spaceskip=0pt\relax}
\providecommand{\BIBentryALTinterwordstretchfactor}{4}
\providecommand{\BIBentryALTinterwordspacing}{\spaceskip=\fontdimen2\font plus
\BIBentryALTinterwordstretchfactor\fontdimen3\font minus \fontdimen4\font\relax}
\providecommand{\BIBforeignlanguage}[2]{{%
\expandafter\ifx\csname l@#1\endcsname\relax
\typeout{** WARNING: IEEEtran.bst: No hyphenation pattern has been}%
\typeout{** loaded for the language `#1'. Using the pattern for}%
\typeout{** the default language instead.}%
\else
\language=\csname l@#1\endcsname
\fi
#2}}
\providecommand{\BIBdecl}{\relax}
\BIBdecl

\bibitem{bari2021real}Bari, B., Islam, M., Rashid, M., Hasan, M., Razman, M., Musa, R., Ab Nasir, A. \& Majeed, A. A real-time approach of diagnosing rice leaf disease using deep learning-based faster R-CNN framework. {\em PeerJ Computer Science}. \textbf{7} pp. e432 (2021)
\bibitem{christou2004potential}Christou, P. \& Twyman, R. The potential of genetically enhanced plants to address food insecurity. {\em Nutrition Research Reviews}. \textbf{17}, 23-42 (2004)
\bibitem{dordas2008role}Dordas, C. Role of nutrients in controlling plant diseases in sustainable agriculture. A review. {\em Agronomy For Sustainable Development}. \textbf{28} pp. 33-46 (2008)
\bibitem{daniya2022deep}Daniya, T. \& Vigneshwari, S. Deep neural network for disease detection in rice plant using the texture and deep features. {\em The Computer Journal}. \textbf{65}, 1812-1825 (2022)

\bibitem{m1}Raha, Avi Deb, et al. "Modeling and Predictive Analytics of Breast Cancer Using Ensemble Learning Techniques: An Explainable Artificial Intelligence Approach." Computers, Materials \& Continua 81.3 (2024).

\bibitem{sharma2022big}Sharma, R. \& Singh, A. Big bang–big crunch-CNN: an optimized approach towards rice crop protection and disease detection. {\em Archives Of Phytopathology And Plant Protection}. \textbf{55}, 143-161 (2022)

\bibitem{m2}Dihan, Fatema Jannat, et al. "MpoxSLDNet: A Novel CNN Model for Detecting Monkeypox Lesions and Performance Comparison with Pre-trained Models." arXiv preprint arXiv:2405.21016 (2024).

\bibitem{liu2023analysis}Liu, H., Cui, Y., Wang, J. \& Yu, H. Analysis and research on rice disease identification method based on deep learning. {\em Sustainability}. \textbf{15}, 9321 (2023)


\bibitem{liang2019rice}Liang, W., Zhang, H., Zhang, G. \& Cao, H. Rice blast disease recognition using a deep convolutional neural network. {\em Scientific Reports}. \textbf{9}, 1-10 (2019)
\bibitem{jiang2023rice}Jiang, M., Feng, C., Fang, X., Huang, Q., Zhang, C. \& Shi, X. Rice disease identification method based on attention mechanism and deep dense network. {\em Electronics}. \textbf{12}, 508 (2023)

\bibitem{m3}Muzahid, Abu Jafar Md, et al. "Optimal safety planning and driving decision-making for multiple autonomous vehicles: A learning based approach." 2021 Emerging Technology in Computing, Communication and Electronics (ETCCE). IEEE, 2021.

\bibitem{feng2020investigation}Feng, L., Wu, B., Zhu, S., Wang, J., Su, Z., Liu, F., He, Y. \& Zhang, C. Investigation on data fusion of multisource spectral data for rice leaf diseases identification using machine learning methods. {\em Frontiers In Plant Science}. \textbf{11} pp. 577063 (2020)
\bibitem{ganesan2022hybridization}Ganesan, G. \& Chinnappan, J. Hybridization of ResNet with YOLO classifier for automated paddy leaf disease recognition: An optimized model. {\em Journal Of Field Robotics}. \textbf{39}, 1085-1109 (2022)
\bibitem{verma2021prediction}Verma, T. \& Dubey, S. Prediction of diseased rice plant using video processing and LSTM-simple recurrent neural network with comparative study. {\em Multimedia Tools And Applications}. \textbf{80}, 29267-29298 (2021)
\bibitem{quach2023using}Quach, L., Quynh, A., Quoc, K. \& Thai, N. Using optimization algorithm to improve the accuracy of the CNN model on the rice leaf disease dataset. {\em Information Systems For Intelligent Systems: Proceedings Of ISBM 2022}. pp. 535-544 (2023)
\bibitem{al2023using}Al-Gaashani, M., Samee, N., Alnashwan, R., Khayyat, M. \& Muthanna, M. Using a Resnet50 with a kernel attention mechanism for rice disease diagnosis. {\em Life}. \textbf{13}, 1277 (2023)
\bibitem{goluguri2021rice}Goluguri, N., Devi, K. \& Srinivasan, P. Rice-net: an efficient artificial fish swarm optimization applied deep convolutional neural network model for identifying the Oryza sativa diseases. {\em Neural Computing And Applications}. \textbf{33}, 5869-5884 (2021)
\bibitem{kukana2020hybrid}Kukana, P. \& Others Hybrid Machine Learning Algorithm-Based Paddy Leave Disease Detection System. {\em 2020 International Conference On Smart Electronics And Communication (ICOSEC)}. pp. 512-519 (2020)
\bibitem{leonard2014importance}Leonard, A. \& Gianessi, P. Importance of Pesticides for Growing Rice in South and South East. {\em International Pesticides Benefit Case Study}. \textbf{108} pp. 30-33 (2014)
\bibitem{singh2019sheath}Singh, P., Mazumdar, P., Harikrishna, J. \& Babu, S. Sheath blight of rice: a review and identification of priorities for future research. {\em Planta}. \textbf{250} pp. 1387-1407 (2019)

\bibitem{m4}Murad, Saydul Akbar, et al. "Computer-aided system for extending the performance of diabetes analysis and prediction." 2021 International Conference on Software Engineering \& Computer Systems and 4th International Conference on Computational Science and Information Management (ICSECS-ICOCSIM). IEEE, 2021.

\bibitem{liu2020first}Liu, L., Zhao, Y., Zhang, Y., Wang, L., Hou, Y. \& Huang, S. First report of leaf spot disease on rice caused by Epicoccum sorghinum in China. {\em Plant Disease}. \textbf{104}, 2735-2735 (2020)
\bibitem{zarbafi2019overview}Zarbafi, S. \& Ham, J. An overview of rice QTLs associated with disease resistance to three major rice diseases: blast, sheath blight, and bacterial panicle blight. {\em Agronomy}. \textbf{9}, 177 (2019)
\bibitem{han2014quantitative}Han, X., Yang, Y., Wang, X., Zhou, J., Zhang, W., Yu, C., Cheng, C., Cheng, Y., Yan, C. \& Chen, J. Quantitative trait loci mapping for bacterial blight resistance in rice using bulked segregant analysis. {\em International Journal Of Molecular Sciences}. \textbf{15}, 11847-11861 (2014)
\bibitem{guo2010identification}Guo, S., Zhang, D. \& Lin, X. Identification and mapping of a novel bacterial blight resistance gene Xa35 (t) originated from Oryza minuta. {\em Scientia Agricultura Sinica}. \textbf{43}, 2611-2618 (2010)
\bibitem{barbedo2016review}Barbedo, J. A review on the main challenges in automatic plant disease identification based on visible range images. {\em Biosystems Engineering}. \textbf{144} pp. 52-60 (2016)
\bibitem{goralski2020artificial}Goralski, M. \& Tan, T. Artificial intelligence and sustainable development. {\em The International Journal Of Management Education}. \textbf{18}, 100330 (2020)
\bibitem{wu2018development}Wu, Y. \& Feng, J. Development and application of artificial neural network. {\em Wireless Personal Communications}. \textbf{102} pp. 1645-1656 (2018)
\bibitem{kujawa2021artificial}Kujawa, S. \& Niedbała, G. Artificial neural networks in agriculture. {\em Agriculture}. \textbf{11} pp. 497 (2021)
\bibitem{escamilla2020applications}Escamilla-García, A., Soto-Zarazúa, G., Toledano-Ayala, M., Rivas-Araiza, E. \& Gastélum-Barrios, A. Applications of artificial neural networks in greenhouse technology and overview for smart agriculture development. {\em Applied Sciences}. \textbf{10}, 3835 (2020)

\bibitem{samborska2014artificial}Samborska, I., Alexandrov, V., Sieczko, L., Kornatowska, B., Goltsev, V., Cetner, M. \& Kalaji, H. Artificial neural networks and their application in biological and agricultural research. {\em J. NanoPhotoBioSciences}. \textbf{2} pp. 14-30 (2014)

\bibitem{m6}Murad, Saydul Akbar, et al. "AI powered asthma prediction towards treatment formulation: An android app approach." Intelligent Automation \& Soft Computing 34.1 (2022): 87-103.


\bibitem{zorzetto2000processing}Zorzetto, L., Maciel Filho, R. \& Wolf-Maciel, M. Processing modelling development through artificial neural networks and hybrid models. {\em Computers  Chemical Engineering}. \textbf{24}, 1355-1360 (2000)
\bibitem{smrekar2009development}Smrekar, J., Assadi, M., Fast, M., Kuštrin, I. \& De, S. Development of artificial neural network model for a coal-fired boiler using real plant data. {\em Energy}. \textbf{34}, 144-152 (2009)
\bibitem{verma2017optimizing}Verma, T. \& Dubey, S. Optimizing rice plant diseases recognition in image processing and decision tree based model. {\em International Conference On Next Generation Computing Technologies}. pp. 733-751 (2017)
\bibitem{azim2021effective}Azim, M., Islam, M., Rahman, M. \& Jahan, F. An effective feature extraction method for rice leaf disease classification. {\em Telkomnika (Telecommunication Computing Electronics And Control)}. \textbf{19}, 463-470 (2021)
\bibitem{yao2009application}Yao, Q., Guan, Z., Zhou, Y., Tang, J., Hu, Y. \& Yang, B. Application of support vector machine for detecting rice diseases using shape and color texture features. {\em 2009 International Conference On Engineering Computation}. pp. 79-83 (2009)
\bibitem{islam2019wavelet}Islam, S. \& Mazumder, B. Wavelet based feature extraction for rice plant disease detection and classification. {\em 2019 3rd International Conference On Electrical, Computer \& Telecommunication Engineering (ICECTE)}. pp. 53-56 (2019)
\bibitem{ghyar2017computer}Ghyar, B. \& Birajdar, G. Computer vision based approach to detect rice leaf diseases using texture and color descriptors. {\em 2017 International Conference On Inventive Computing And Informatics (ICICI)}. pp. 1074-1078 (2017)
\bibitem{saputra2020rice}Saputra, R., Wasiyanti, S., Saefudin, D., Supriyatna, A., Wibowo, A. \& Others Rice leaf disease image classifications using KNN based on GLCM feature extraction. {\em Journal Of Physics: Conference Series}. \textbf{1641}, 012080 (2020)

\bibitem{m5}Islam, Md Shamimul, et al. "A deep Spatio-temporal network for vision-based sexual harassment detection." 2021 Emerging Technology in Computing, Communication and Electronics (ETCCE). IEEE, 2021.

\bibitem{matin2020efficient}Matin, M., Khatun, A., Moazzam, M. \& Uddin, M. An efficient disease detection technique of rice leaf using AlexNet. {\em Journal Of Computer And Communications}. \textbf{8}, 49-57 (2020)
\bibitem{lu2017identification}Lu, Y., Yi, S., Zeng, N., Liu, Y. \& Zhang, Y. Identification of rice diseases using deep convolutional neural networks. {\em Neurocomputing}. \textbf{267} pp. 378-384 (2017)
\bibitem{rahman2020identification}Rahman, C., Arko, P., Ali, M., Khan, M., Apon, S., Nowrin, F. \& Wasif, A. Identification and recognition of rice diseases and pests using convolutional neural networks. {\em Biosystems Engineering}. \textbf{194} pp. 112-120 (2020)
\bibitem{latif2022deep}Latif, G., Abdelhamid, S., Mallouhy, R., Alghazo, J. \& Kazimi, Z. Deep learning utilization in agriculture: Detection of rice plant diseases using an improved CNN model. {\em Plants}. \textbf{11}, 2230 (2022)
\bibitem{simhadri2023automatic}Simhadri, C. \& Kondaveeti, H. Automatic recognition of rice leaf diseases using transfer learning. {\em Agronomy}. \textbf{13}, 961 (2023)
\bibitem{rice_leaf_dataset}VBookshelf Rice Leaf Diseases Dataset. (https://www.kaggle.com/datasets/vbookshelf/rice-leaf-diseases,2023), Accessed: 17 January 2023
\bibitem{narin2021automatic}Narin, A., Kaya, C. \& Pamuk, Z. Automatic detection of coronavirus disease (covid-19) using x-ray images and deep convolutional neural networks. {\em Pattern Analysis And Applications}. \textbf{24} pp. 1207-1220 (2021)
\bibitem{narin2021automatic}Narin, A., Kaya, C. \& Pamuk, Z. Automatic detection of coronavirus disease (covid-19) using x-ray images and deep convolutional neural networks. {\em Pattern Analysis And Applications}. \textbf{24} pp. 1207-1220 (2021)
\bibitem{chen2020epidemiological}Chen, N., Zhou, M., Dong, X., Qu, J., Gong, F., Han, Y., Qiu, Y., Wang, J., Liu, Y., Wei, Y. \& Others Epidemiological and clinical characteristics of 99 cases of 2019 novel coronavirus pneumonia in Wuhan, China: a descriptive study. {\em The Lancet}. \textbf{395}, 507-513 (2020)
\bibitem{bijoy2024towards}Bijoy, M., Hasan, N., Biswas, M., Mazumdar, S., Jimenez, A., Ahmed, F., Rasheduzzaman, M. \& Momen, S. Towards Sustainable Agriculture: A Novel Approach for Rice Leaf Disease Detection Using dCNN and Enhanced Dataset. {\em IEEE Access}. (2024)
\bibitem{pizer1987adaptive}Pizer, S., Amburn, E., Austin, J., Cromartie, R., Geselowitz, A., Greer, T., Haar Romeny, B., Zimmerman, J. \& Zuiderveld, K. Adaptive histogram equalization and its variations. {\em Computer Vision, Graphics, And Image Processing}. \textbf{39}, 355-368 (1987)
\bibitem{he1987texture}He, D., Wang, L. \& Guibert, J. Texture feature extraction. {\em Pattern Recognition Letters}. \textbf{6}, 269-273 (1987)
\bibitem{de2013multi}De Siqueira, F., Schwartz, W. \& Pedrini, H. Multi-scale gray level co-occurrence matrices for texture description. {\em Neurocomputing}. \textbf{120} pp. 336-345 (2013)
\bibitem{sen2009counting}Sen, G., Liu, W. \& Yan, H. Counting people in crowd open scene based on grey level dependence matrix. {\em 2009 International Conference On Information And Automation}. pp. 228-231 (2009)
\bibitem{duhamel1990fast}Duhamel, P. \& Vetterli, M. Fast Fourier transforms: a tutorial review and a state of the art. {\em Signal Processing}. \textbf{19}, 259-299 (1990)
\bibitem{nason1994discrete}Nason, G. \& Silverman, B. The discrete wavelet transform in S. {\em Journal Of Computational And Graphical Statistics}. \textbf{3}, 163-191 (1994)
\bibitem{abdi2010principal}Abdi, H. \& Williams, L. Principal component analysis. {\em Wiley Interdisciplinary Reviews: Computational Statistics}. \textbf{2}, 433-459 (2010)
\bibitem{scholkopf1997kernel}Schölkopf, B., Smola, A. \& Müller, K. Kernel principal component analysis. {\em International Conference On Artificial Neural Networks}. pp. 583-588 (1997)
\bibitem{makhzani2013k}Makhzani, A. \& Frey, B. K-sparse autoencoders. {\em ArXiv Preprint ArXiv:1312.5663}. (2013)
\bibitem{zabalza2016novel}Zabalza, J., Ren, J., Zheng, J., Zhao, H., Qing, C., Yang, Z., Du, P. \& Marshall, S. Novel segmented stacked autoencoder for effective dimensionality reduction and feature extraction in hyperspectral imaging. {\em Neurocomputing}. \textbf{185} pp. 1-10 (2016)
\bibitem{archer1997sensitivity}Archer, G., Saltelli, A. \& Sobol, I. Sensitivity measures, ANOVA-like techniques and the use of bootstrap. {\em Journal Of Statistical Computation And Simulation}. \textbf{58}, 99-120 (1997)
\bibitem{tallarida1987chi}Tallarida, R., Murray, R., Tallarida, R. \& Murray, R. Chi-square test. {\em Manual Of Pharmacologic Calculations: With Computer Programs}. pp. 140-142 (1987)
\bibitem{belgiu2016random}Belgiu, M. \& Drăguţ, L. Random forest in remote sensing: A review of applications and future directions. {\em ISPRS Journal Of Photogrammetry And Remote Sensing}. \textbf{114} pp. 24-31 (2016)
\bibitem{gupta1997artificial}Gupta, N., VAYA, N. \& MISHRA, B. Artificial Neural Network. {\em Eastern Pharmacist}. \textbf{40}, 39-41 (1997)
\bibitem{wang2022review}Wang, J., Lu, S., Wang, S. \& Zhang, Y. A review on extreme learning machine. {\em Multimedia Tools And Applications}. \textbf{81}, 41611-41660 (2022)
\bibitem{kathiresan2021disease}Kathiresan, G., Anirudh, M., Nagharjun, M. \& Karthik, R. Disease detection in rice leaves using transfer learning techniques. {\em Journal Of Physics: Conference Series}. \textbf{1911}, 012004 (2021)
\bibitem{daniya2022exponential}Daniya, T. \& Vigneshwari, S. Exponential Rider-Henry Gas Solubility optimization-based deep learning for rice plant disease detection. {\em International Journal Of Information Technology}. 


\end{thebibliography}

\end{document}